\documentclass[runningheads]{llncs}
\usepackage{amsmath}
\usepackage{comment}
\usepackage[T1]{fontenc}
\usepackage{cite}
\usepackage{hyperref}
\usepackage{graphicx}
\usepackage{float}
\usepackage{array}
\usepackage{arydshln} 
\usepackage{caption}
\usepackage[T1]{fontenc}
\usepackage{pgfplots}
\usepackage{subcaption}
\pgfplotsset{compat=1.17}
\usepackage{graphicx,verbatim}
\begin{document}
\title{Toward Markerless Video-based Tremor Analysis: \\Objective Quantification of Pathological Tremor \\in Mouse Preclinical Models}
\titlerunning{Markerless Video-based Tremor Analysis}

\author{Yota Koshimoto\inst{1} \and
Akihiro Tsukahara\inst{1} \and \\
Yasuhiro Moriwaki\inst{1} \and
Mariko Isogawa\inst{1}}

\authorrunning{Y. Koshimoto et al.}

\institute{Keio University, Yokohama, Japan\\
\email{\{koshimoto.yota, mariko.isogawa\}@keio.jp}}
  
\maketitle           
\begin{abstract}
Tremor is a movement disorder characterized by involuntary, rhythmic oscillations of body parts and is a hallmark of several neurological conditions, including Parkinson’s disease and essential tremor. Elucidating its underlying mechanisms relies heavily on mouse models, which offer genetic manipulability and translational relevance to human neural circuitry. Accordingly, these models are indispensable for studying tremor pathophysiology. So far, electromyography and accelerometers have been used as methods to quantitatively observe tremors in mice.
However, these methods have several drawbacks, such as high costs and complex setups. In particular, the invasive surgical implantation of devices causes significant stress to the animals. Although RGB-based methods offer non-invasive and cost-effective alternatives, they often lack the sensitivity required to detect subtle tremors. Therefore, this paper addresses these challenges by achieving mouse tremor severity estimation using conventional RGB cameras only. To address the challenging task of isolating tremor-related vibrations while the mouse itself is also in motion, our pipeline incorporates segmentation-based preprocessing to extract the mouse region and a Tremor Score Estimation Module that captures subtle tremors with high sensitivity. In the experiments, we assessed tremors in unrestrained mice using a non-invasive method with two standard cameras. The results demonstrated a strong correlation with accelerometer measurements and confirmed that the method accurately captured the intensity-dependent characteristics of tremors. The project page is available at \url{https://isogawalab.github.io/Video-based-Tremor-Analysis-Project/}.


\end{abstract}
\section{Introduction}
Parkinson’s disease (PD) has become an increasingly significant global health concern as the aging population continues to expand, underscoring the urgent need for effective therapeutic strategies. Tremor, defined as involuntary and rhythmic oscillatory movement, is one of the cardinal motor symptoms of PD. To investigate its underlying mechanisms and evaluate potential treatments, genetically manipulable PD model mice are widely used to reproduce key pathological and behavioral features of the disease. Accordingly, precise and objective quantification of murine tremor is essential for accurately assessing therapeutic efficacy and facilitating the translation of preclinical findings into clinical applications.

However, conventional tremor assessment relies on visual observation by human evaluators~\cite{zhang_gut_2023, JOLICOEUR1991317}, leading to issues of subjectivity and inconsistency. Furthermore, such manual evaluation methods are time-consuming and suffer from poor reproducibility. To ensure objectivity, previous studies have employed invasive, contact-based approaches such as accelerometers~\cite{HALLBERG1985261, park, kistler}, electromyography sensors~\cite{Bekar2008, Hosoi6339, Gunther1983}, and marker-based motion capture systems~\cite{Ignatowska-JankowskaENEURO.0045-25.2025}. While quantitative, these methods restrict natural movement and may compromise behavioral observations.

To address this issue, alternative non-invasive methods using piezoelectric film sensors~\cite{AJIMA2021109074} and force plate-based systems~\cite{fowler_force-plate_2001, https://doi.org/10.1002/btm2.10432} have been proposed. However, these solutions require specialized equipment and high costs. 
Using inexpensive and widely used conventional RGB cameras is one possible option. However, the existing implementation~\cite{https://doi.org/10.1002/btm2.10432} has two major limitations in tremor detection. First, because the method depends solely on movement velocity, it lacks the sensitivity to detect the subtle oscillations that characterize tremors. Second, its reliance on a top-down perspective makes it difficult to capture tremors occurring in the anti-gravity (vertical) direction. Collectively, these constraints compromise the overall precision of tremor quantification. 

To overcome these limitations, this paper proposes a framework for estimating mouse tremor severity using only RGB videos. Specifically, our framework extracts body part positions from an animal pose estimation model using a horizontal camera perspective. We incorporate a Move Detection Module to isolate involuntary oscillations from voluntary locomotion and prevent the false detection of movement as a tremor. Additionally, we propose a Tremor Score Estimation Module specifically to sensitively quantify microscopic oscillations into a tremor score. By focusing on movement along the vertical axis, our approach is uniquely capable of capturing tremors occurring in the anti-gravity direction, which were previously undetectable from top-down views.
This quantitative assessment of subtle tremors facilitates the investigation of the relationship between drug dosage and tremor severity, thereby aiding in the development of therapeutic strategies for Parkinson’s disease.

In summary, our contributions are as follows.
\begin{itemize}
    \item We propose a novel, non-invasive framework that quantifies tremor severity with high precision using only side-view RGB videos.
    \item We construct an original dataset of  RGB videos synchronized with accelerometer data, enabling rigorous quantitative evaluation of video-based tremor assessment.
    \item We propose two novel modules: a Move Detection Module that distinguishes voluntary locomotion from involuntary tremor oscillations, and a Tremor Score Estimation Module that directly quantifies wide-range tremor severity from coordinates of a body part.
\end{itemize}

\section{Related Work}
\subsection{Sensor Based Mouse Tremor Measurement}
\subsubsection{Invasive Methods.}
 To quantitatively assess mouse tremors, various methods have been proposed. Sensor-based approaches utilizing accelerometers~\cite{HALLBERG1985261, park, kistler} or electromyography~\cite{Bekar2008, Hosoi6339, Gunther1983} require surgical implantation of devices. While these methods provide objective physiological data, they necessitate animal restraint and device attachment, which disrupt natural behavior.

In recent years, a marker-based motion capture system has emerged as an innovative approach for tremor analysis in mice~\cite{Ignatowska-JankowskaENEURO.0045-25.2025}. This system involves surgical implantation of multiple reflective markers beneath the skin, with subsequent 3D motion tracking using synchronized high-speed cameras. This multi-sensor implantation system enables comprehensive visualization of whole-body tremor dynamics, allowing for precise anatomical localization of tremor activity. However, this approach retains significant limitations: first, the necessity of specialized surgical instrumentation, and second, the inherently invasive nature of subcutaneous sensor implantation procedures in mouse models.

\subsubsection{Non-Invasive Methods.}
Acknowledging the inherent limitations of conventional tremor assessment, a study incorporating piezoelectric film-based tremor measurement with advanced image analysis methods~\cite{AJIMA2021109074} was conducted. It presents a quantitative evaluation of microtremors in unrestrained mouse subjects. However, this method requires specialized equipment, such as a piezoelectric sensor.

Recent studies employ markerless pose estimation to analyze tremor dynamics through overhead video capture~\cite{https://doi.org/10.1002/btm2.10432}. While this approach attempts to quantify tremor using RGB data, it faces two fundamental limitations. First, the method relies on inter-frame movement velocity, which lacks the sensitivity to distinguish the subtle oscillations characteristic of tremors from voluntary motion. Second, the exclusive use of a top-down perspective constrains motion capture predominantly to the horizontal plane. Because tremor severity is often reflected in oscillatory movements during anti-gravity postural maintenance, insufficient representation of vertical components may reduce measurement accuracy.

While our method shares the objective of previous research to achieve objective and quantitative tremor assessment, it addresses the previously unexplored problem of estimating anti-gravity tremor severity using side-view video data acquired from RGB cameras. It allows tremor measurements to be made in a simple filming environment consisting of two conventional cameras, with minimal interference with the subject's natural behavior and no reliance on subjective observer assessment.

\subsection{Video-based Animal Behavior Analysis with Pose Estimation}
The advent of deep learning-based markerless pose estimation frameworks, particularly DeepLabCut (DLC)~\cite{deeplabcut} and SLEAP~\cite{sleap}, has significantly advanced quantitative behavioral analysis in mouse models. These models enable automated tracking of anatomical keypoints from standard video, facilitating objective assessment of motor phenotypes without physical instrumentation.

Recent applications demonstrate the utility of pose estimation for quantifying motor deficits in neurological disease models. For instance, skeletal keypoints in Parkinson's disease models have been tracked using DLC~\cite{ossigood2024AutomatedPT}, enabling objective classification of motor behaviors and eliminating manual scoring subjectivity. Similarly, unsupervised clustering of pose dynamics~\cite{b-soid_2021} successfully distinguished L-DOPA-induced dyskinesia from normal grooming behaviors. Furthermore, locomotor ataxia has been quantified by analyzing spatial variability and inter-joint coordination~\cite{machado2020}, revealing disease-specific gait signatures.

In this work, we extend the application of Animal Pose Estimation to the domain of video-based tremor analysis. Accordingly, this work proposes a novel tremor assessment method leveraging animal pose estimation models to capture tremor dynamics in a simple recording environment without restricting natural behavior, enabling the demonstration of tremor severity estimation from RGB videos.

\section{Method}
\begin{figure}[t]
    \includegraphics[width=\textwidth]{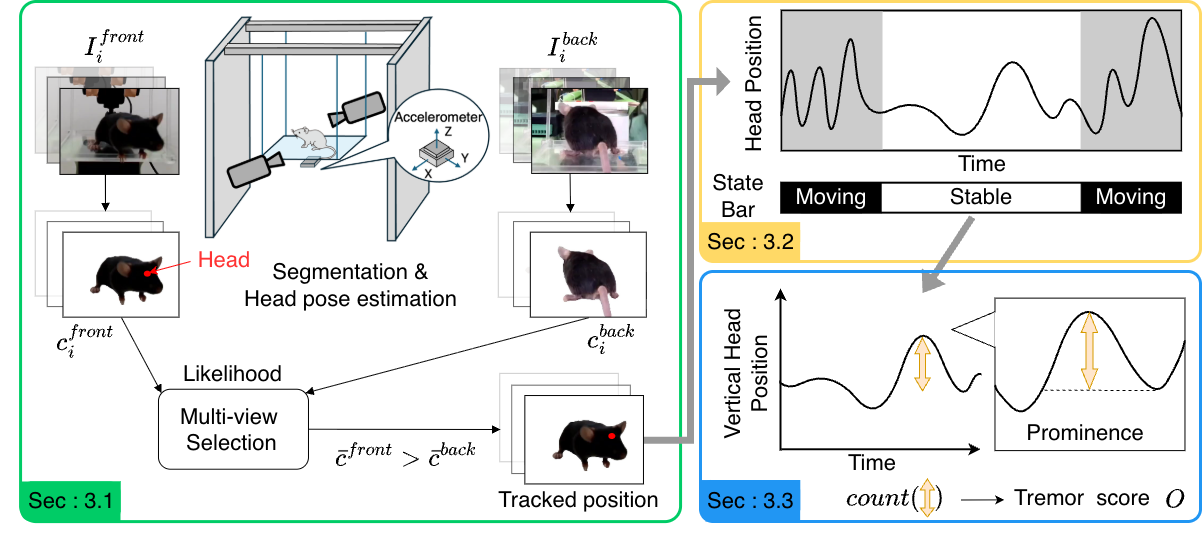}
    \caption{Overview of the proposed framework.
    } 
    \label{fig1}
\end{figure}

Figure~\ref{fig1} illustrates our framework. Given dual-view video sequences $\mathbf{I}^{\mathrm{front}}$ and $\mathbf{I}^{\mathrm{back}}$ of a freely 
moving mouse, our goal is to estimate tremor severity $O$ as a scalar score. Our framework consists of three components: Sec.~\ref{3-1} Multi-view Selection to choose the optimal viewing angle, Sec.~\ref{3-2} Move Detection Module to separate locomotion from tremor, and Sec.~\ref{3-3} Tremor Score Estimation Module to quantify tremor intensity. We detail each component below.

\subsection{Multi-view Selection Strategy}
\label{3-1}
To ensure robust tracking regardless of the mouse's orientation, our system captures video from two horizontal perspectives. Since the mouse frequently changes direction, causing self-occlusion, one view typically provides clearer visibility of the target body parts than the other. First, to isolate the subject from the complex background, we employ SAM 2~\cite{ravi2024sam} for segmentation-based preprocessing, generating a segmentation mask of the mouse in each frame  $\mathbf{I}_i$. From the masked region, the head position $(x_i, y_i)$ and its associated tracking confidence $c_i$ are extracted using an animal pose estimation model. The head is selected as the primary tracking target due to its prominent tremor expression and lower occlusion risk compared to limbs. Subsequently, the trajectory is partitioned into $M$ non-overlapping segments, denoted as $\{V_m\}_{m=1}^{M}$, each with a fixed duration of $T$ seconds. For each segment, the view with the higher mean confidence $\bar{c}_m$ is selected for subsequent analysis.

\subsection{Move Detection Module}
\label{3-2}
Distinguishing involuntary tremor from voluntary locomotion is crucial for accurate assessment. To address this, we propose a module that filters out intervals dominated by active movement using the trajectory selected in Sec.~\ref{3-1}. Let the trajectory of the selected view for the segment $V_m$ be denoted as a sequence of coordinates $\mathbf{P}_m = \{(x_0, y_0), (x_1, y_1), \dots, (x_{N-1}, y_{N-1})\}$, where $N$ is the number of frames within the segment duration $T$. We calculate the cumulative movement distance $d_m$ as the sum of Euclidean distances between consecutive frames.

If $d_m$ is below a predefined threshold, the segment is identified as a stationary state and is preserved for tremor scoring. Conversely, segments with excessive movement are discarded to prevent locomotion artifacts.
In this study, we empirically set the segment duration $T = 3$ seconds (corresponding to $N=90$ frames at 30 fps) and the movement threshold to 400 pixels, corresponding to approximately 80 mm using our pixel-to-millimeter calibration factor of 0.2 mm per pixel estimated from held-out pilot data.

\subsection{Tremor Score Estimation Module}\label{3-3}
This module quantifies tremor severity within segments identified as stationary. The primary challenge in tremor quantification is to suppress non-pathological noise while sensitively capturing subtle tremors. To achieve this, we employ peak prominence analysis rather than absolute amplitude, focusing on the vertical (Y-axis) displacement.
Let $\mathbf{Y}_m = \{y_0, y_1, \dots, y_{N-1}\}$ be the Y-coordinate sequence for a segment $V_m$. For a local maximum at index $p$, we define its left and right reference boundaries, $L(p)$ and $R(p)$, as the nearest indices where the signal height equals or exceeds $y_p$, or the sequence endpoints:
\begin{equation}
    \begin{aligned}
    L(p) &= \max \{ i < p \mid y_i \ge y_p \text{ or } i = 0 \} \\
    R(p) &= \min \{ i > p \mid y_i \ge y_p \text{ or } i = N-1 \}
    \end{aligned}
\end{equation}
The prominence $P(p)$ is then determined by the vertical distance from the peak to the higher of the two local minima (bases) within these boundaries:
\begin{equation}
    P(p) = y_p - \max \left( \min_{i \in [L(p), p]} y_i, \;\min_{i \in [p, R(p)]} y_i \right)
\end{equation}
Unlike absolute peak height, this metric considers the relative depth of surrounding valleys, providing a more robust measure of oscillatory intensity by filtering out minor positional fluctuations. 
The global tremor score $O$ for the entire observation period is then calculated as the total count of peaks whose prominence exceeds a predefined threshold $\tau$. In this study, we set $\tau=1.0$ pixel, corresponding to approximately 0.2 mm according to our calibration factor.

\section{Experimental Settings}
\textbf{Dataset.}
We collected an original dataset comprising 56 sequences of RGB videos (paired front/back views of 28 wild-type mice). We used wild-type mice divided into four groups based on harmaline dosage: 0 mg/kg (saline, n=6), 5 mg/kg (n=8), 10 mg/kg (n=7) and 20 mg/kg (n=7). Mice were placed on a custom-built suspension platform. As the onset of prominent tremors was typically observed at 2.5 minutes, we analyzed the subsequent 5-minute interval to ensure the evaluation of the active tremor phase. Videos were recorded at a resolution of 1280 $\times$ 720 pixels at 30 fps. We calibrated the pixel-to-millimeter scale using a checkerboard with a known square size placed near the center of the platform. Both front- and back-view cameras were positioned at the same distance and configured with the same image resolution, yielding a local scale of 0.2 mm per pixel at the measurement plane. This scale can be re-estimated for different experimental setups using the same calibration procedure. To obtain ground-truth motion data, we utilized a wireless 3-axis accelerometer (MVP-RF8-S-V170, MicroStone Co., Ltd.) attached to the suspension platform.

\noindent
\textbf{Evaluation Metric.}
Following established accelerometer-based tremor assessment protocols~\cite{carlsen_accurate_2019}, we used the peak Power Spectral Density (PSD) of the accelerometer signal as the reference measure of tremor intensity. We evaluated the proposed video-based tremor score by computing its correlation with the accelerometer-derived PSD.

\noindent
\textbf{Baseline Method.}
To evaluate the performance of our framework, we compared our method with two baselines:
(1) Ni et al.~\cite{https://doi.org/10.1002/btm2.10432}: we employed a prior approach that estimates mouse tremor intensity from RGB video. This method involves tracking the animal's center in overhead video via a pose estimation model and deriving a score from the peak value of the movement's Power Spectral Density (PSD). To ensure a fair comparison using our dataset, we applied this pipeline to a side-view perspective.
(2) Human-subjective scoring: A manual assessment was conducted based on the clinical scoring criteria established in Zhang et al.\cite{zhang_gut_2023}. For each analyzed segment, an experienced evaluator scored the tremor severity, and the total sum of these scores across the measurement period was calculated to represent the subjective tremor intensity.

\noindent
\textbf{Implementation Details.}
For pose estimation, we employed DeepLabCut with a ResNet-50 backbone~\cite{he2015deepresiduallearningimage}. The model was trained on 480 frames (20 frames manually annotated from each of the 24 videos) to ensure robust tracking.

\section{Experiments and Results}
To verify the effectiveness of our framework, we conducted comparative experiments using the peak Power Spectral Density (PSD) from accelerometer data as the ground-truth for tremor intensity.

\begin{table}[t]
    \centering
    \caption{Comparison of correlation coefficient}
    \label{tab:correlations}
    \begin{tabular}[t]{lcc@{\hskip 1mm}lccc@{\hskip -1mm}ccc@{\hskip 1mm}ccc@{\hskip 1mm}ccc@{\hskip 1mm}ccc}
        \hline
        Methods&   Correlation coefficient \\
        \hline
        Ni et al.~\cite{https://doi.org/10.1002/btm2.10432} & -0.27 \\
        Human-subjective evaluation & 0.82\\
        \textbf{Ours} & \textbf{0.86}\\
        \hdashline
        Ours w/o multi-view strategy & 0.82\\
        Ours w/o seg. preprocessing & 0.75\\
        Ours w/o Move Detection Module& 0.78\\
        \hline
    \end{tabular}
\end{table}

\begin{figure}[t]
  \centering
  \begin{minipage}{0.32\textwidth}
    \centering
    \includegraphics[width=\linewidth]{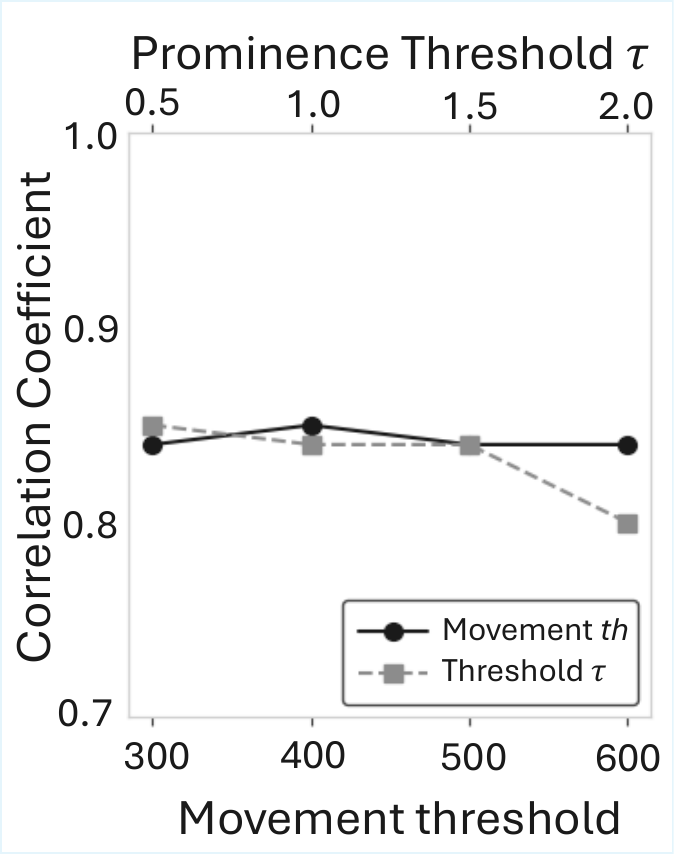}
    \caption{Sensitivity analysis of hyperparameters.}
    \label{fig:4-4}
  \end{minipage}
    \hfill
  \begin{minipage}{0.32\textwidth}
    \centering
    \includegraphics[width=\linewidth]{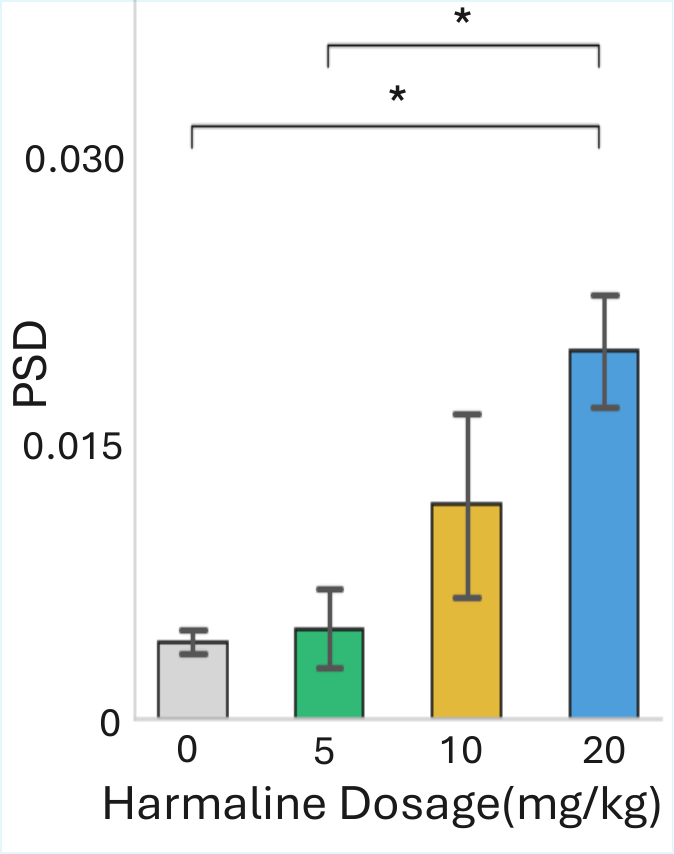}
    \caption{Relation between dosage and PSD.}
    \label{fig:4-1}
  \end{minipage}
  \hfill
  \begin{minipage}{0.32\textwidth}
    \centering
    \includegraphics[width=\linewidth]{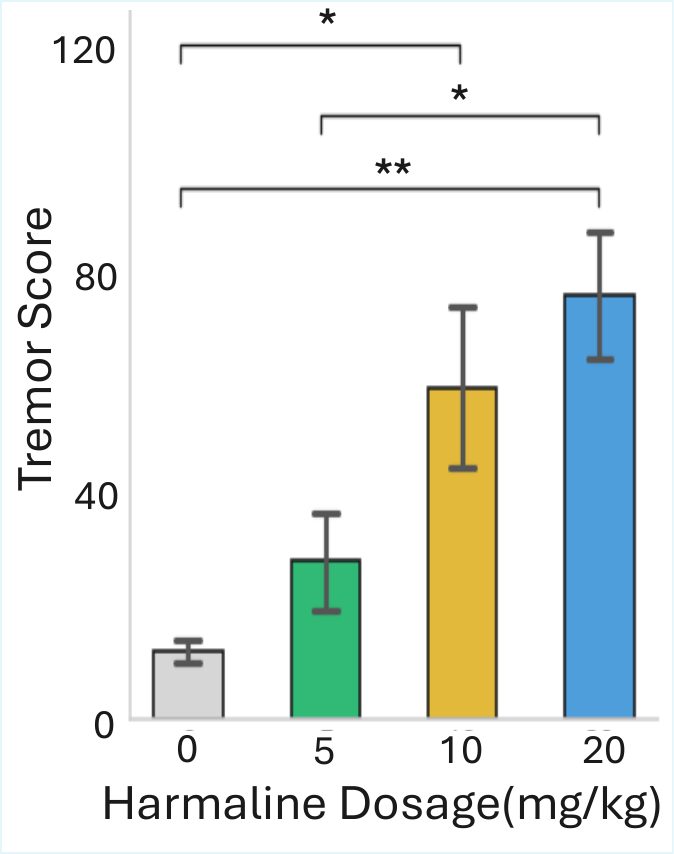}
    \caption{Relation between dosage and tremor score.}
    \label{fig:4-2}
  \end{minipage}
\end{figure}

\begin{figure}[t]
    \centering
    \includegraphics[width=1.0\hsize]{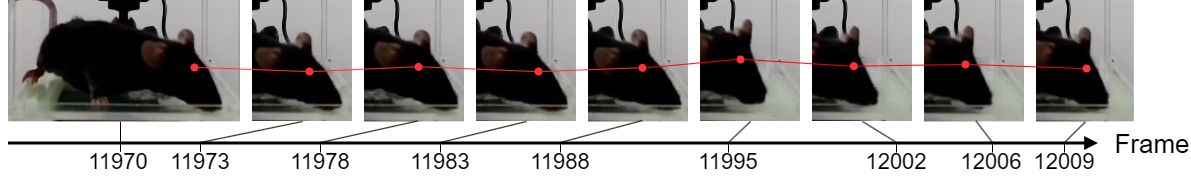}
    \caption{An example head displacement during tremor.}
    \label{fig:4-5} 
\end{figure}

\noindent\textbf{Comparison with Baseline Methods.} We compared our method with a velocity-based automated baseline~\cite{https://doi.org/10.1002/btm2.10432} and human-subjective scoring. In Table~\ref{tab:correlations}, the RGB baseline showed a negative correlation ($r = -0.27$) with the accelerometer ground truth, as simple cumulative displacement is easily dominated by locomotive noise and lacks sensitivity to subtle tremors. In contrast, while human-subjective scoring achieved a high correlation ($r = 0.82$), it remains inherently limited by its discrete, coarse nature and the prohibitive time-cost of manual frame-by-frame analysis, making it impractical for large-scale studies. 

\noindent\textbf{Ablation Study.} We evaluated the contribution of each module to the estimation accuracy (Table~\ref{tab:correlations}). The performance declined when key components were removed, with the correlation coefficient dropping to $r=0.75$ without the segmentation-based preprocessing and to $r=0.78$ without the Move Detection Module. These results underscore the necessity of isolating the mouse region and filtering out locomotion artifacts to capture subtle tremor dynamics. Additionally, excluding the multi-view selection strategy resulted in a decrease to $r=0.82$, confirming its role in maintaining robust tracking against self-occlusion.
 
\noindent\textbf{Sensitivity Check on Hyperparameters.} We assessed the framework's robustness against hyperparameter variations. As shown in Fig.~\ref{fig:4-4}, the correlation coefficient remained stable above $0.80$ across various settings for both the movement threshold and prominence threshold $\tau$. This stability indicates that our method is not overly sensitive to specific parameter tuning, ensuring reliable performance across different environments and tremor intensities.

\noindent\textbf{Pharmacological Validation.} 
In animal studies for Parkinson's disease research, tremor severity under pharmacological manipulation is a key indicator for evaluating drug efficacy and disease mechanisms. Therefore, we evaluated the practical utility of our framework by replicating the physiological dose-response relationship of harmaline. We investigated whether the tremor severity scores estimated by our method increase with dosage during harmaline administration, where tremor severity is known to exhibit a dose-dependent increase~\cite{AJIMA2021109074}. A one-way ANOVA was performed to examine the effect of harmaline dosage (saline, 5, 10 and 20 mg/kg) on tremor severity, revealing a significant main effect ($p < 0.05$). As shown in Fig.~\ref{fig:4-1}, post-hoc tests confirmed significantly higher tremor intensity in the 10 and 20 mg/kg groups. Our vision-based tremor scores (Fig.~\ref{fig:4-2}) exhibited a highly concordant trend, correctly capturing the dose-dependent increase. These results indicate that our non-invasive approach serves as a reliable alternative to invasive sensors for drug efficacy evaluation.

\noindent\textbf{Anti-Gravity (Vertical) Tremor.} As shown in Fig.~\ref{fig:4-5}, the mouse exhibits rhythmic vertical head oscillations during tremors. These anti-gravity movements are inherently difficult to observe from a conventional top-down view. Thus, our side-view configuration is essential to explicitly visualize these vertical displacements and comprehensively evaluate tremor severity.

\section{Conclusion}
This study proposes a non-invasive tremor measurement framework using standard RGB cameras. Leveraging mouse pose estimation and prominence-based peak detection, the method extracts subtle tremor oscillations while suppressing behavioral noise. Results demonstrate a strong correlation with accelerometer-based ground truth ($r = 0.86$), achieving accuracy comparable to expert manual scoring while eliminating its time burden. Furthermore, successfully replicating dose-dependent tremor patterns highlights the system’s utility for pharmacological evaluation. Our framework overcomes the limitations of contact-based sensors and subjective observation, providing a scalable platform to accelerate drug discovery and mechanistic research. Notably, its ability to analyze tremors during resting states underscores its broad translational potential, extending its applicability to Parkinsonian tremor and other pathological movement disorders.

\vspace{1em}
\noindent
\textbf{{Ethics Approval Statement.}}
This study was conducted with the approval of the Ethics Committee of Keio University (Approval No. A2022-341).

\noindent\textbf{Acknowledgements.}
This work was partially supported by the KGRI Challenge Grant, JSPS KAKENHI Grant Number 25H01159, and the grant from the Smoking Research Foundation.

\noindent
\textbf{Disclosure of Interests.}
The authors have no competing interests to declare that are relevant to the content of this article.

\bibliographystyle{splncs04}
\bibliography{ref}

\end{document}